\documentclass[pdflatex,sn-mathphys-num]{sn-jnl}% Math and Physical Sciences Numbered Reference Style
\usepackage{graphicx}%
\usepackage{multirow}%
\usepackage{amsmath,amssymb,amsfonts}%
\usepackage{amsthm}%
\usepackage{mathrsfs}%
\usepackage[title]{appendix}%
\usepackage{xcolor}%
\usepackage{textcomp}%
\usepackage{manyfoot}%
\usepackage{booktabs}%
\usepackage{algorithm}%
\usepackage{algorithmicx}%
\usepackage{algpseudocode}%
\usepackage{listings}%
\usepackage{makecell}%
\usepackage{pifont}%
\newcommand{\cmark}{\textcolor{green!60!black}{\ding{51}}}%
\theoremstyle{thmstyleone}%
\theoremstyle{thmstyletwo}%

\theoremstyle{thmstylethree}%

\begin{document}

% \title[Article Title]{Article Title}
\title{SO3UFormer: Learning Intrinsic Spherical Features for Rotation-Robust Panoramic Dense Prediction} 


% \author*[1,2]{\fnm{First} \sur{Author}}\email{iauthor@gmail.com}



% \affil*[1]{\orgdiv{Department}, \orgname{Organization}, \orgaddress{\street{Street}, \city{City}, \postcode{100190}, \state{State}, \country{Country}}}

% \affil[2]{\orgdiv{Department}, \orgname{Organization}, \orgaddress{\street{Street}, \city{City}, \postcode{10587}, \state{State}, \country{Country}}}

% \affil[3]{\orgdiv{Department}, \orgname{Organization}, \orgaddress{\street{Street}, \city{City}, \postcode{610101}, \state{State}, \country{Country}}}
\author[1,2]{\fnm{Qinfeng} \sur{Zhu}}

\author[3]{\fnm{Yunxi} \sur{Jiang}}

\author*[2]{\fnm{Lei} \sur{Fan}}\email{lei.fan@xjtlu.edu.cn}

\affil[1]{\orgname{University of Liverpool}, \orgaddress{\country{United Kingdom}}}

\affil[2]{\orgname{Xi'an Jiaotong-Liverpool University}, \orgaddress{\country{China}}}

\affil[3]{\orgname{CNRS}, \orgaddress{\country{France}}}


\abstract{Panoramic dense-prediction models, spanning semantic segmentation and depth estimation, are typically trained under a strict gravity-aligned assumption. Real-world captures, however, routinely violate it: handheld devices jitter and aerial platforms change attitude, so the camera is rarely upright. Under such 3D reorientation, standard spherical Transformers overfit global latitude cues and collapse. We introduce SO3UFormer, an architecture that learns intrinsic spherical features largely decoupled from the underlying coordinate frame, through three geometric components: (1) removing absolute latitude encoding, which breaks the dependence on the gravity axis; (2) quadrature-consistent spherical attention, which corrects for non-uniform sampling density; and (3) a gauge-aware relative positional bias built from local tangent-plane angles rather than global axes. A logit-space \emph{SO(3)}-consistency regularizer, used only during training, further suppresses residual discretization effects. To benchmark robustness, we introduce Pose35, a variant of Stanford2D3D perturbed by random rotations within $\pm 35^\circ$, and evaluate under a full, arbitrary \emph{SO(3)} stress test. There, the baseline SphereUFormer collapses from 67.53 \emph{mIoU} on Pose35 to 25.26 under the full \emph{SO(3)} test, whereas SO3UFormer reaches 72.03 on Pose35 and retains 70.67 under the same test. Similarly, on a second real-world dataset (Matterport3D) for segmentation and on panoramic depth estimation, SO3UFormer remains essentially rotation-invariant while the gravity-anchored baseline again loses most of its accuracy. Code and models are available at https://github.com/zhuqinfeng1999/SO3UFormer.}




\maketitle

\section{Introduction}
\label{sec:intro}

Panoramic imaging offers a comprehensive $360^\circ$ field of view \cite{gao2022review,elharrouss2021panoptic,zhang2024taming}, serving as a critical modality for embodied agents \cite{arif2024panoramic}, aerial drones \cite{zheng2024efficient}, and immersive reality applications \cite{ullah2022perceptual}. Unlike standard perspective cameras \cite{deng2009imagenet}, panoramic sensors capture the entire surrounding geometry in a single shot \cite{xu2025motion}. However, this completeness comes with a hidden assumption in current computer vision pipelines: \textit{canonical gravity alignment} \cite{bergmann2025anchor}. Most state-of-the-art panoramic dense-prediction models, from semantic segmentation to depth estimation \cite{benny2025sphereuformer}, implicitly assume the camera is perfectly upright. While this assumption often holds for street-view datasets collected by specialized vehicles, it breaks down in dynamic capture settings such as aerial and mobile robotic platforms \cite{sun2021aerial}. Drones tilt during flight, handheld devices jitter, and robots move over uneven terrain, leading to significant roll and pitch rotations \cite{wu2025digital}.

\begin{figure}[t]
  \centering
  \includegraphics[width=\linewidth]{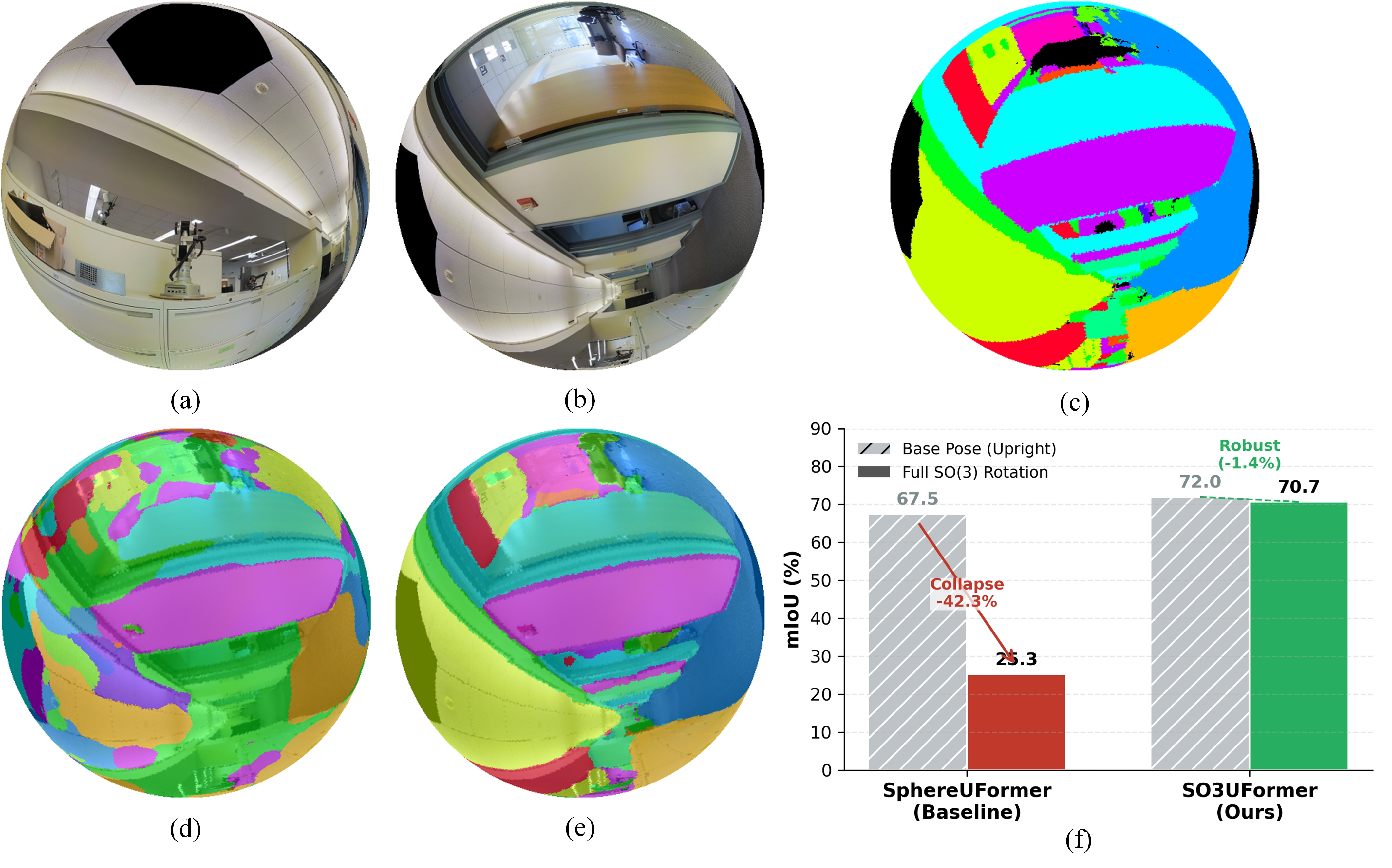}
    \caption{\textbf{Breaking the Gravity Lock: Rotation Robustness in Panoramic Segmentation.} 
    We present a comparison between a canonical upright view (a) and the same scene under an arbitrary \textbf{\emph{SO(3)} rotation} (b), mimicking real-world unconstrained motion. 
    (c) shows the Ground Truth semantic map for the rotated input. 
    (d) The state-of-the-art \textbf{SphereUFormer} \cite{benny2025sphereuformer} fails catastrophically on the rotated input, as it relies on absolute latitude cues (gravity bias) and cannot recognize the tilted geometry. 
    (e) In contrast, our \textbf{SO3UFormer} reduces reliance on global coordinate cues and produces a consistent segmentation that closely matches the ground truth. 
    (f) Quantitative results confirm that while the baseline performance collapses by over 42\% \emph{mIoU} under rotation, our method maintains robust accuracy (70.7\%), effectively closing the \emph{SO(3)} domain gap.}
    \label{fig:teaser}
\end{figure}

The consequence of this assumption is a fragility to 3D rotations that is often masked by standard benchmarks \cite{chen20233d}. Existing methods, whether based on Equirectangular Projection (ERP) \cite{jung2025edm} or  \cite{candela2024object}, typically bake in absolute positional encodings (e.g., latitude and longitude) \cite{shen2022panoformer}. These encodings act as a strong prior, anchoring the semantic understanding to a global ``North Pole.'' When the camera rotates, the physical ``floor'' is no longer at the bottom of the image, yet the model, misguided by absolute coordinates, continues to look for it there. As we demonstrate in this work, this reliance causes a \textit{catastrophic performance collapse} (Figure~\ref{fig:teaser}): a leading spherical Transformer, SphereUFormer \cite{benny2025sphereuformer}, drops from \textbf{67.53\%} mean Intersection-over-Union (\emph{mIoU}) to \textbf{25.26\%} under arbitrary \emph{SO(3)} rotations \cite{makinen2008rotation}. The model effectively learns to segment the \textit{coordinate system}, not the scene geometry \cite{sahin2025non}.

To address this fundamental limitation, we propose to shift the paradigm from \textit{extrinsic coordinate learning} to \textit{intrinsic geometric perception}. We introduce \textbf{SO3UFormer}, a rotation-robust spherical Transformer designed to be robust to changes in the underlying coordinate frame. Our approach is built on the insight that robustness requires respecting the intrinsic geometry of the sphere \cite{cao2024geometric}, rather than overfitting dataset-specific global orientation cues.

SO3UFormer implements this philosophy through three architectural imperatives. First, we eliminate \textbf{absolute latitude encoding}, removing the global ``gravity bias'' that misleads the network during rotation. Second, we introduce Quadrature-Consistent Attention. Since spherical grids (like the icosahedral subdivision) do not have uniform area measures, standard attention mechanisms can bias aggregation towards denser sampling regions. We correct this by integrating quadrature weights into the attention normalization. Third, we develop a Gauge-Aware Relative Positional Mechanism. Instead of relying on global axes, we define relative geometry using local tangent-plane projected angles and a discrete gauge pooling scheme over a small set of in-plane frame rotations, so that positional reasoning does not depend on a privileged global reference. Furthermore, we use geometry-consistent down/up sampling across scales and a training-time logit-space \emph{SO(3)}-consistency regularizer based on the same index-based spherical resampling scheme.

To rigorously evaluate robustness beyond standard benchmarks, we introduce \textbf{Pose35}, a stress-test protocol based on Stanford2D3D \cite{armeni2017joint2d3d} with random pose perturbations ($\pm 35^\circ$), and evaluate with a \textit{full, arbitrary} \emph{SO(3)} stress test. Under this extreme setting, SO3UFormer demonstrates remarkable stability, achieving \textbf{72.03\%} \emph{mIoU} on Pose35 and retaining \textbf{70.67\%} \emph{mIoU} under full \emph{SO(3)} rotations, substantially reducing the large performance gap left by prior methods. We develop and analyze SO3UFormer primarily through semantic segmentation, but its operators are defined purely from spherical geometry and therefore apply to panoramic dense prediction in general; we confirm this by transferring the same model, unchanged, to depth estimation.

Our main contributions are summarized as follows:
\begin{itemize}
    \item We identify the root cause of rotation fragility in panoramic dense prediction: the coupling of absolute coordinate embeddings with measure-inconsistent aggregation.
    \item We propose \textbf{SO3UFormer}, a rotation-robust spherical Transformer that improves robustness by combining gauge-aware relative geometry, quadrature-consistent spherical attention, and the removal of gravity-dependent biases.
    \item We design a suite of \emph{SO(3)}-friendly operators, including geometry-consistent multi-scale down/up sampling, reinforced by a rotation-consistency regularizer.
    \item We establish a new benchmark protocol using the \textbf{Pose35} dataset. Extensive experiments show that SO3UFormer dramatically improves robustness under full 3D rotations, reducing catastrophic failure and achieving 70.67 \emph{mIoU} in the \emph{SO(3)} out-of-distribution (OOD) stress test.
    \item We demonstrate that the same intrinsic design generalizes beyond the primary benchmark, reproducing the collapse-versus-stability pattern on a second real-world dataset (Matterport3D) and on panoramic depth estimation, indicating that the rotation robustness stems from the geometric formulation rather than from any single task or dataset.
\end{itemize}

\section{Related Work}
\label{sec:relat}

\subsection{Panoramic Dense Prediction}
Semantic segmentation for 2D images has matured and has been widely deployed in domains such as remote sensing and medical imaging \cite{zhu2025classwise, zhang2025advances}. Many advances in panoramic dense prediction inherit this progress \cite{zheng2024open}. Classical fully convolutional designs and their successors, including FCN, U-Net, and deep residual backbones, established a strong foundation for dense prediction in the planar setting \cite{long2015fcn,ronneberger2015unet,he2016resnet}. Multi-scale encoder--decoder paradigms further improved global context modeling and boundary localization, exemplified by PSPNet and DeepLabv3+ \cite{zhao2017pspnet,chen2018deeplabv3plus}. More recently, masked attention and Transformer-based segmentation have become mainstream, offering flexible long-range aggregation and strong scalability \cite{dosovitskiy2021vit,liu2021swin,xie2021segformer,cheng2022mask2former}.

A core challenge for $360^\circ$ panoramas is the ERP, which introduces latitude-dependent distortion and non-uniform sampling density \cite{yoon2022spheresr,shidanshidi2013non}. Early attempts addressed this issue by adapting convolutional operators to spherical geometry or by learning directly on spherical representations for recognition and dense prediction \cite{su2017sphericalconv,coors2018spherenet}. Following the same geometric principle, recent panoramic dense-prediction models increasingly shift from ERP-specific heuristics to intrinsic spherical feature learning. Monocular depth estimation is the other central dense-prediction task on panoramas, where methods such as PanoFormer \cite{shen2022panoformer} and Elite360D \cite{ai2024elite360d} mitigate distortion through tangent-patch and bi-projection representations. Unifying both tasks, SphereUFormer builds a spherical U-shaped Transformer for dense scene understanding, spanning segmentation and depth, over $360^\circ$ signals \cite{benny2025sphereuformer}. In indoor environments, 2D--3D semantic correspondence datasets with Stanford2D3D-style annotations enable controlled and reproducible comparisons under consistent scene semantics \cite{armeni2017joint2d3d}.

\subsection{Deep Learning on Spherical Surfaces}
Applying standard planar convolution kernels directly on ERP images is problematic near the poles, where severe distortion can corrupt local appearance and confuse the model \cite{eder2020tangent}. Learning directly on the sphere avoids ERP singularities and allows operators to respect the manifold structure \cite{coors2018spherenet,perraudin2019deepsphere,ai2022deep}. Representative work leverages group structure on $\mathrm{SO}(3)$ or spherical harmonics to construct spherical convolutional networks, including spherical CNNs with provable equivariance properties \cite{cohen2018sphericalcnns}. Beyond strictly spherical convolution, group-equivariant and steerable designs generalize convolution to broader transformation groups and provide strong inductive biases that often improve sample efficiency and robustness \cite{cohen2016groupconv,weiler2019e2cnn}.

Gauge equivariance further clarifies how to treat local reference frames (tangent frames) when defining orientation-dependent features on curved surfaces. Gauge-equivariant convolutional networks formalize this viewpoint and show how to build consistent local operators while controlling frame ambiguity \cite{cohen2019gauge}. In particular, the Icosahedral CNN instantiation demonstrates that gauge-equivariant convolutions on an icosahedral approximation of the sphere can be implemented efficiently with standard \texttt{conv2d} primitives while remaining effective for omnidirectional segmentation \cite{cohen2019gauge}. These ideas align with the broader goals of geometric deep learning, which advocates architectures whose symmetries match those of the data domain \cite{bronstein2017gdl}.

\subsection{Rotation Equivariance and Geometric Attention}
Attention mechanisms \cite{vaswani2017attention} provide a flexible means of long-range aggregation, but naively applying attention on non-Euclidean domains can couple representations to arbitrary coordinate choices. This motivates equivariant formulations that respect underlying symmetries. In 3D, Tensor Field Networks provide a principled framework for rotation/translation-equivariant feature interactions \cite{thomas2018tfn}, while SE(3)-Transformers extend these ideas to attention-like message passing that preserves equivariance \cite{fuchs2020se3transformer}. Related graph-based approaches, such as E(n)-equivariant GNNs, offer practical and scalable equivariant updates for geometric learning \cite{satorras2021egnn}. More recently, Bonev et al. derives spherical attention directly on $\mathbb{S}^2$ and shows that quadrature-weighted discretization and geodesic neighborhood attention can improve geometric fidelity and approximate rotational equivariance for Transformer-style models on spherical signals \cite{bonev2025attention}.

Closer to our setting, a parallel line treats robustness to 3D camera disturbance directly on panoramas without enforcing exact equivariance: SGAT4PASS introduces spherical geometry-aware projection and deformable patch embedding with a panorama-aware loss, improving stability of panoramic segmentation under small pitch and roll perturbations~\cite{li2023sgat4pass}.
Our work shares this geometry-aware, non-strictly-equivariant philosophy but targets the full $\mathrm{SO}(3)$ regime and attributes robustness to a combination of intrinsic local operators rather than projection-level corrections alone.

Collectively, these developments suggest two recurring requirements for rotation-robust dense prediction: avoiding hard-coding absolute global coordinates \cite{liu2024visual}; and ensuring that aggregation is consistent with the geometry and symmetries of the underlying domain. Our work follows this line by combining spherical attention, gauge-aware relative geometry, and $\mathrm{SO}(3)$-consistent multi-scale processing for panoramic dense prediction.

\section{Methodology}
\label{sec:metho}

\subsection{Preliminaries and Strategy}
\noindent\textbf{Spherical signal and icosahedral discretization.}
A panoramic observation is naturally a signal on the unit sphere \cite{liu2023uav},
\( \mathbf{x}: \mathbb{S}^2 \rightarrow \mathbb{R}^{C} \).
Following prior spherical transformers, we discretize \(\mathbb{S}^2\) using an icosahedral subdivision.
Let \( \{\mathbf{p}_i\}_{i=1}^{L} \subset \mathbb{S}^2 \) denote mesh nodes (vertices or face normals depending on the node type),
each carrying a feature vector \( \mathbf{x}_i \in \mathbb{R}^{D} \).
Mesh connectivity defines a local neighborhood \(\mathcal{N}(i)\), and all attention in this work is computed locally on \(\mathcal{N}(i)\).

\noindent\textbf{Area weights used in our implementation.}
Each node is associated with a positive area weight \(\omega_i>0\), estimated from the icosphere geometry \cite{suliman2022geomorph}.
In our implementation, \(\omega\) is \emph{mean-normalized}:
\begin{equation}
\bar{\omega} \;=\;\frac{1}{L}\sum_{i=1}^{L}\omega_i,\qquad
\omega_i \leftarrow \frac{\omega_i}{\bar{\omega}},
\label{eq:omega_mean_norm}
\end{equation}
so that \(\frac{1}{L}\sum_i \omega_i = 1\).
This normalization keeps the magnitude of the quadrature correction numerically stable across ranks.

\noindent\textbf{Design strategy: reducing dependence on a privileged global axis.}
Gravity-aligned benchmarks encourage shortcut learning from absolute latitude cues and chart-dependent angular offsets \cite{pintore2021deep3dlayout}.
When roll--pitch changes occur (e.g., tilting during turns or handheld jitter), these shortcuts become unreliable and can fail badly \cite{qin2018vins,cadena2017past}.
We therefore aim to learn \emph{intrinsic} spherical features by: (1) removing absolute latitude encoding,
(2) correcting attention aggregation using spherical area weights,
(3) replacing chart-indexed relative bias with a gauge-pooled angular encoding defined in local tangent planes, and
(4) regularizing predictions to be consistent under the same spherical resampling mechanism used during training.

\noindent\textbf{Approximate equivariance via intrinsic locality.}
We state our claim precisely. We do \emph{not} assert exact $\mathrm{SO}(3)$ equivariance in the sense of spherical-harmonic or steerable convolutional networks~\cite{cohen2018sphericalcnns,cohen2019gauge,weiler2019e2cnn}.
Instead, we pursue \emph{approximate equivariance through intrinsic locality}: on the continuous sphere $\mathbb{S}^2$, every operator we introduce is defined purely from intrinsic quantities, namely geodesic distances, tangent-plane angles, and area measures, and never references a global axis.
Such operators are invariant under the $\mathrm{SO}(3)$ action by construction, because a global rotation transports a node together with its entire intrinsic neighborhood, leaving all pairwise geodesic and tangent-plane relations unchanged.
Exact invariance is broken only by the icosahedral discretization: a finite mesh does not admit a continuous rotation group, an arbitrary rotation maps mesh nodes to off-lattice positions that must be resampled, and the vertices possess merely a five-fold ($C_5$) symmetry rather than continuous rotational symmetry.
The residual discrepancy introduced by this discretization is exactly what the training-time $\mathrm{SO}(3)$-consistency regularizer (Sec.~\ref{sec:reg}) is designed to suppress.
This places SO3UFormer between two extremes: unlike absolute-coordinate Transformers it removes the gravity-aligned shortcut, and unlike strictly equivariant convolutional networks it retains the representational capacity of unconstrained attention, trading exact equivariance for a favorable accuracy--robustness operating point that we quantify in Sec.~\ref{sec:exper}.

\subsection{Overall Architecture}
\begin{figure}[t]
  \centering
  \includegraphics[width=\linewidth]{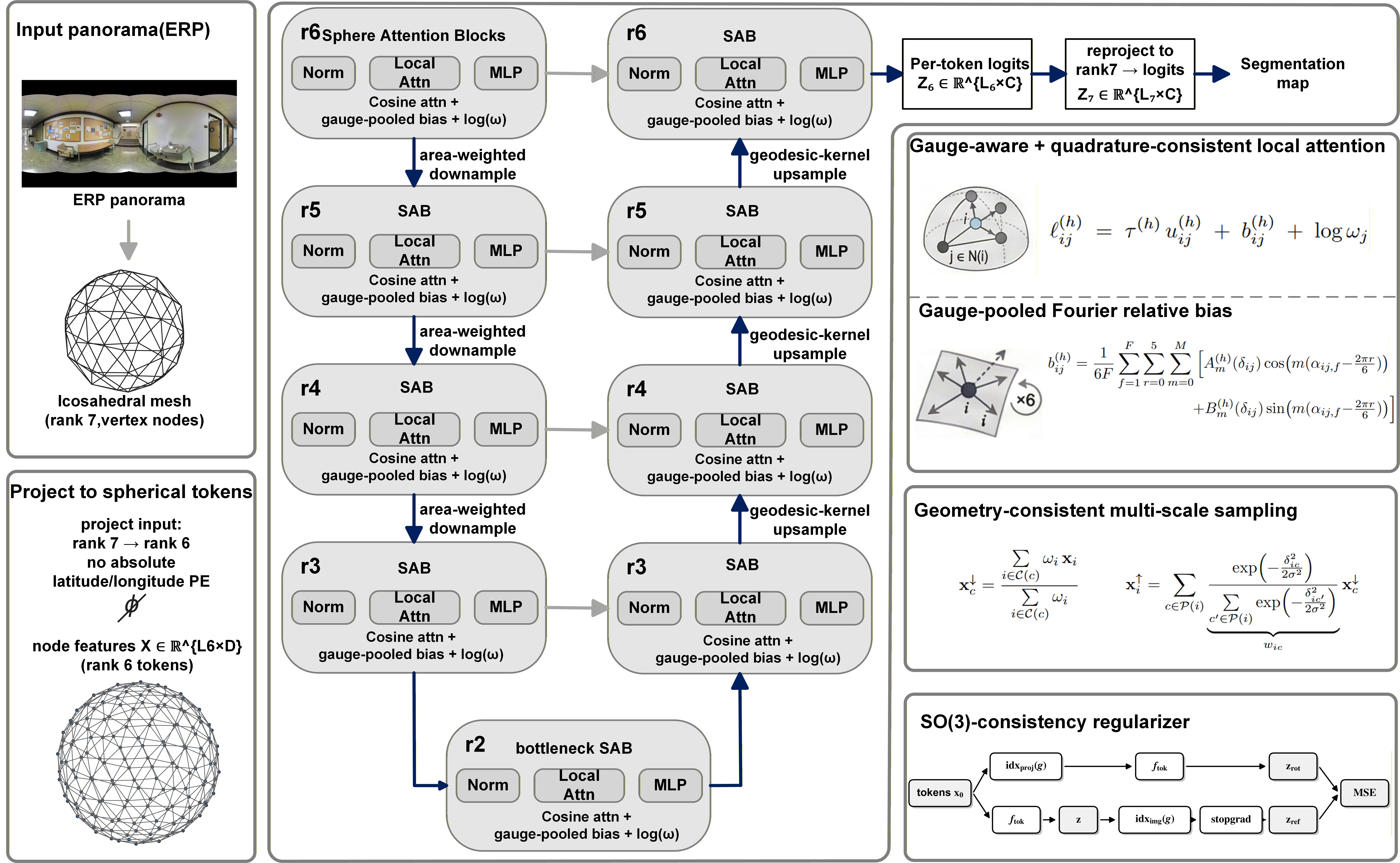}
  \caption{\textbf{SO3UFormer overview.} A U-shaped spherical Transformer with gauge-aware, quadrature-consistent local attention,
  geometry-consistent down/up sampling, and an \emph{SO(3)}-consistency regularizer via spherical index-based resampling.}
  \label{fig:method_overview}
\end{figure}

\noindent\textbf{Backbone.}
\textbf{SO3UFormer} follows a U-shaped multi-scale spherical Transformer backbone (Figure~\ref{fig:method_overview}) with an encoder--decoder over multiple icosphere ranks
and symmetric skip connections.
The input panorama is first projected to spherical tokens; the U-shaped backbone operates on the projected token ranks,
and predictions are finally lifted back to the output node set.
Each stage contains several attention blocks composed of normalization, local spherical self-attention, and an MLP.
Our changes are localized to: (1) attention logits and positional bias (Sec.~\ref{sec:gauge_attn}),
(2) geometry-consistent down/up sampling (Sec.~\ref{sec:sampling}),
and (3) an \emph{SO(3)}-consistency regularizer (Sec.~\ref{sec:reg}).

\noindent\textbf{Design rationale and complementarity.}
The proposed modifications address different failure sources that become visible under 3D camera reorientation.
Removing absolute latitude encoding suppresses the most direct shortcut to gravity-aligned semantics.
Quadrature-consistent local attention and the gauge-pooled Fourier relative positional bias then replace this shortcut with local interactions that are defined by spherical geometry: the former reduces aggregation bias induced by non-uniform sampling density, while the latter encodes relative angular structure in local tangent planes without introducing a global reference axis.
Geometry-consistent down/up sampling extends the same principle across scales, so that the multi-scale pathway does not reintroduce chart-dependent distortions through interpolation.
Finally, the \emph{SO(3)}-consistency regularizer is used only during training and encourages predictions to remain consistent under the same index-based spherical resampling used in our pipeline, which helps reduce sensitivity to discretization and resampling artifacts rather than changing the inference procedure itself.

\begin{table}[t]
\centering
\caption{Operator-level comparison of SphereUFormer and SO3UFormer, with the $\mathrm{SO}(3)$ property of each SO3UFormer operator on the \emph{continuous} sphere. Our design replaces every chart- or topology-dependent operation with an intrinsic one; exact invariance holds in the continuous limit and is approximated on the icosahedral mesh.}
\label{tab:operator_equivariance}
\small
\setlength{\tabcolsep}{4pt}
\renewcommand{\arraystretch}{1.25}
\begin{tabular}{llll}
\toprule
Operator & SphereUFormer & SO3UFormer (ours) & Continuous-$\mathbb{S}^2$ property \\
\midrule
Absolute latitude PE & used & removed & sole global-axis term \\
Attention aggregation & uniform softmax & $+\log\omega_j$ quadrature & area-consistent, invariant \\
Relative positional bias & $(\Delta\theta,\Delta\phi)$ chart & tangent-plane $\alpha$, $6$-pool, $F$-anchor & gauge-invariant \\
Downsampling & nearest parent & area-weighted scatter & intrinsic, invariant \\
Upsampling & bilinear pair & geodesic Gaussian kernel & intrinsic, invariant \\
Training regularizer & none & $\mathrm{SO}(3)$-consistency MSE & soft toward equivariance \\
\bottomrule
\end{tabular}
\end{table}

\noindent\textbf{Reading the operator table.}
Table~\ref{tab:operator_equivariance} also previews why the components contribute unequally (Sec.~\ref{sec:ablation}): the absolute-latitude encoding is the only operator explicitly tied to a global axis, so removing it yields the single largest gain under reorientation, whereas the remaining rows replace chart-dependent or measure-inconsistent computations with intrinsic ones that are invariant on the continuous sphere and exact only up to discretization. The same property explains why these operators also help upright accuracy.

\noindent\textbf{Rank flow in the spherical backbone.}
The spherical hierarchy is defined on projected spherical tokens, not directly on the ERP grid.
In our setup, the input panorama is first sampled on rank-7 output/image support and then projected to \emph{rank-6} spherical tokens for backbone processing.
The internal U-shaped hierarchy operates on these tokens with the rank flow
\(r6 \rightarrow r5 \rightarrow r4 \rightarrow r3\), a bottleneck at \(r2\), and then upsampling back to \(r6\).
Per-token logits are predicted on rank-6 tokens and are finally reprojected to the rank-7 output sampling for evaluation and visualization.
This separation keeps the internal multi-scale computation on spherical tokens while handling input/output sampling through dedicated projection and reprojection operators.

\noindent\textbf{Notation.}
At an internal spherical-token scale \(s\), let \(\mathbf{X}^{(s)}\in\mathbb{R}^{L_s\times D}\) denote node features, where \(L_s\) is the number of nodes at that rank.
When the scale is clear from context, we write \(L\) for brevity.
We use \(L_{\mathrm{img}}\) specifically for the final image/output sampling (rank-7 in our setup).
For query node \(i\), keys/values are gathered from \(\mathcal{N}(i)\) (with padding and a mask for batching).
We use multi-head attention with \(H\) heads.

\subsection{Gauge-Aware Spherical Attention}
\label{sec:gauge_attn}

\subsubsection{Quadrature-consistent cosine attention}
\noindent\textbf{Cosine-similarity logits with a learnable scale.}
For head \(h\), we compute linear projections and \(\ell_2\)-normalize queries and keys:
\begin{equation}
\mathbf{q}_i^{(h)} = \mathrm{norm}\!\left(\mathbf{W}_Q^{(h)} \mathbf{x}_i\right),\quad
\mathbf{k}_j^{(h)} = \mathrm{norm}\!\left(\mathbf{W}_K^{(h)} \mathbf{x}_j\right),\quad
\mathbf{v}_j^{(h)} = \mathbf{W}_V^{(h)} \mathbf{x}_j ,
\end{equation}
so the similarity term is a cosine similarity
\begin{equation}
u_{ij}^{(h)} = \left\langle \mathbf{q}_i^{(h)}, \mathbf{k}_j^{(h)} \right\rangle \in[-1,1].
\end{equation}
We further apply a learned positive scale \(\tau^{(h)}\) with an \emph{upper clamp only}:
\begin{equation}
\tau^{(h)} \;=\; \exp\!\Big(\min\big(s^{(h)},\log(100)\big)\Big),
\label{eq:logit_scale}
\end{equation}
where \(s^{(h)}\) is a learned scalar.
This prevents extremely sharp attention by capping the scale at \(100\) (no lower clamp is used).

\noindent\textbf{Logits with relative bias and quadrature correction.}
For \(j\in\mathcal{N}(i)\), the attention logit is
\begin{equation}
\ell_{ij}^{(h)} \;=\; \tau^{(h)}\,u_{ij}^{(h)} \;+\; b_{ij}^{(h)} \;+\; \log \omega_j,
\label{eq:attn_logit_qlogit}
\end{equation}
where \(b_{ij}^{(h)}\) is a learnable gauge-pooled relative positional bias
and \(\omega_j\) is the mean-normalized area weight in Eq.~\eqref{eq:omega_mean_norm}.
We then apply softmax over the neighborhood:
\begin{equation}
a_{ij}^{(h)} \;=\; \frac{\exp(\ell_{ij}^{(h)})}{\sum_{k\in\mathcal{N}(i)}\exp(\ell_{ik}^{(h)})},
\qquad
\mathbf{y}_i^{(h)} = \sum_{j\in\mathcal{N}(i)} a_{ij}^{(h)}\,\mathbf{v}_j^{(h)}.
\label{eq:attn_out}
\end{equation}

For valid mesh neighbors, \(\omega_j>0\) by construction.
In the implementation, a small positive clamp is applied before taking \(\log \omega_j\) for numerical safety on buffered tensors, and padded entries are masked before softmax; this does not change the valid-neighbor form in Eq.~\eqref{eq:attn_logit_qlogit}.
Intuitively, adding \(\log\omega_j\) makes attention approximate an area-weighted aggregation on the sphere,
reducing bias induced by non-uniform sampling density.

\subsubsection{Gauge-pooled Fourier relative positional bias}
\label{sec:gauge_bias}

\noindent\textbf{Geodesic distance.}
For nodes \(\mathbf{p}_i,\mathbf{p}_j\in\mathbb{S}^2\), we compute the geodesic distance by
\begin{equation}
\delta_{ij}=\arccos(\mathbf{p}_i^\top\mathbf{p}_j)\in[0,\pi].
\end{equation}

\noindent\textbf{Anchor-based tangent-plane angles.}
Our implementation defines relative angles in the tangent plane of the query node using \(F\) anchors per query node \(i\).
In the reported configuration, anchors are selected from \(\mathcal{N}(i)\setminus\{i\}\) by the geometry-based \texttt{tangent\_max} rule: candidates are ranked by the tangent-plane magnitude
\(\|\mathbf{p}_j-(\mathbf{p}_j^\top\mathbf{p}_i)\mathbf{p}_i\|=\sqrt{1-(\mathbf{p}_j^\top\mathbf{p}_i)^2}\),
and the top \(F\) are used (repeating the last selected index if fewer than \(F\) candidates are available).
For each selected anchor \(a_f(i)\) (\(f=1,\dots,F\)), we form tangent-plane directions by projection:
\begin{equation}
\mathbf{t}_{i\leftarrow j} \;=\;
\frac{\mathbf{p}_j-(\mathbf{p}_j^\top\mathbf{p}_i)\mathbf{p}_i}{\left\|\mathbf{p}_j-(\mathbf{p}_j^\top\mathbf{p}_i)\mathbf{p}_i\right\|},
\qquad
\mathbf{t}_{i\leftarrow a_f} \;=\;
\frac{\mathbf{p}_{a_f(i)}-(\mathbf{p}_{a_f(i)}^\top\mathbf{p}_i)\mathbf{p}_i}{\left\|\mathbf{p}_{a_f(i)}-(\mathbf{p}_{a_f(i)}^\top\mathbf{p}_i)\mathbf{p}_i\right\|}.
\end{equation}
Let \(\mathbf{t}^{\perp}_{i\leftarrow a_f} = \mathbf{p}_i \times \mathbf{t}_{i\leftarrow a_f}\) denote a \(90^\circ\) in-plane rotation.
We define the relative angle (in \((-\pi,\pi]\)) via
\begin{equation}
\alpha_{ij,f} \;=\; \mathrm{atan2}\!\big(\langle \mathbf{t}_{i\leftarrow j}, \mathbf{t}^{\perp}_{i\leftarrow a_f}\rangle,\; \langle \mathbf{t}_{i\leftarrow j}, \mathbf{t}_{i\leftarrow a_f}\rangle\big).
\label{eq:alpha_anchor}
\end{equation}

In implementation, the projection norms are \(\epsilon\)-clamped for numerical stability when constructing the local frame. This makes the bias depend on \emph{local} angular geometry around \(i\) without referencing global longitude/latitude offsets.

\noindent\textbf{Continuous radial bins via linear interpolation.}
We normalize the geodesic distance as
\begin{equation}
\hat{\delta}_{ij} \;=\; \delta_{ij}/\pi \in [0,1],
\end{equation}
which matches the implementation's stored normalized distance. Distance-dependent coefficients are stored on discrete radial bins, but \(\delta_{ij}\) is mapped to bins continuously and evaluated by linear interpolation.
With \(B\) bins and normalized distance \(\hat{\delta}_{ij}\in[0,1]\),
\begin{equation}
t = \hat{\delta}_{ij}(B-1),\quad
b_0=\lfloor t\rfloor,\quad
b_1=\min(b_0+1,B-1),\quad
\eta=t-b_0,
\end{equation}
and any tabulated coefficient \(C(\delta)\) is evaluated as
\begin{equation}
C(\delta_{ij}) \;=\; (1-\eta)\,C[b_0] + \eta\,C[b_1].
\label{eq:lerp_bins}
\end{equation}

\noindent\textbf{Fourier series with fixed 6-rotation pooling and \(F\)-anchor averaging.}
For each head \(h\), we use a truncated Fourier series of order \(M\) in the angle variable,
with learnable distance-dependent coefficients \(A^{(h)}_m(\delta)\) and \(B^{(h)}_m(\delta)\)
(tabulated on radial bins and interpolated by Eq.~\eqref{eq:lerp_bins}).
We then perform a fixed pooling over \emph{six} in-plane rotations and average over the \(F\) anchors.
The resulting relative bias is
\begin{equation}
b_{ij}^{(h)}=
\frac{1}{6F}\sum_{f=1}^{F}\sum_{r=0}^{5}\sum_{m=0}^{M}
\Big[
A_m^{(h)}(\delta_{ij})\cos\!\big(m(\alpha_{ij,f}-\tfrac{2\pi r}{6})\big)+
B_m^{(h)}(\delta_{ij})\sin\!\big(m(\alpha_{ij,f}-\tfrac{2\pi r}{6})\big)
\Big].
\label{eq:fourier_gauge_pool}
\end{equation}
The \(m=0\) sine contribution is identically zero, but we keep it in the summation to match the implementation-aligned parameterization.
The pre-pooled Fourier basis parameterizes local angular responses in each anchor-defined tangent frame, while the subsequent uniform six-rotation averaging projects the bias onto the rotation-commensurate subspace.

\noindent\textbf{Exact mode selection and invariance to the anchor count.}
The uniform pooling over six in-plane rotations is not heuristic smoothing but an exact projection onto the Fourier modes whose order is a multiple of six. For any integer $m$,
\begin{equation}
\frac{1}{6}\sum_{r=0}^{5}\cos\!\Big(m\big(\alpha-\tfrac{2\pi r}{6}\big)\Big)
=
\begin{cases}
\cos(m\alpha), & m\equiv 0 \!\!\pmod 6,\\[2pt]
0, & \text{otherwise},
\end{cases}
\label{eq:mode_selection}
\end{equation}
and identically for the sine terms (the inner sum is the geometric series $\mathrm{Re}\sum_{r=0}^{5} e^{\mathrm{i}\,2\pi m r/6}$, equal to $6$ when $6\mid m$ and $0$ otherwise).
At $M=2$ only $m=0$ survives, and since $\sin(0)=0$, Eq.~\eqref{eq:fourier_gauge_pool} collapses to
\begin{equation}
b_{ij}^{(h)} \;=\; \frac{1}{F}\sum_{f=1}^{F} A_0^{(h)}(\delta_{ij}) \;=\; A_0^{(h)}(\delta_{ij}),
\label{eq:bias_reduces}
\end{equation}
a per-head, purely radial profile independent of the anchor index $f$ and therefore \emph{provably invariant} to the anchor count $F$. We keep the anchor averaging and the higher-order ($m\ge 1$) coefficients as a structural margin: inert at $M=2$, they let the same parameterization express orientation-selective bias once the truncation order passes the first rotation-commensurate harmonic at $m=6$, with no change to the surrounding architecture.

\subsection{Geometry-Consistent Sampling}
\label{sec:sampling}

\noindent\textbf{Downsampling via cosine-argmax parent assignment.}
Between consecutive ranks, each fine node is assigned to a coarse \emph{parent} by maximizing cosine similarity
between the corresponding unit normals:
\begin{equation}
\mathrm{parent}(i) \;=\; \arg\max_{c\in\mathcal{V}_{\mathrm{coarse}}}\;\mathbf{p}_i^\top \mathbf{p}_c.
\label{eq:parent_argmax}
\end{equation}
This provides an efficient approximate mapping between scales.

\noindent\textbf{Area-weighted downsampling.}
Given children \(\mathcal{C}(c)=\{i:\mathrm{parent}(i)=c\}\), we pool features with area weights:
\begin{equation}
\mathbf{x}^{\downarrow}_c =
\frac{\sum\limits_{i\in\mathcal{C}(c)} \omega_i\,\mathbf{x}_i}{\sum\limits_{i\in\mathcal{C}(c)} \omega_i}.
\label{eq:area_down_impl}
\end{equation}
This prevents regions with denser sampling from dominating the multi-scale aggregation.

\noindent\textbf{Upsampling with local candidates and a geodesic kernel.}
For each fine node \(i\), interpolation candidates are restricted to its parent and the parent's 1-ring neighbors:
\(\mathcal{P}(i)=\{\mathrm{parent}(i)\}\cup \mathcal{N}_{1}\big(\mathrm{parent}(i)\big)\).
We then apply a normalized geodesic kernel:
\begin{equation}
\mathbf{x}^{\uparrow}_i =
\sum_{c\in\mathcal{P}(i)}
\underbrace{
\frac{\exp\!\left(-\frac{\delta_{ic}^2}{2\sigma^2}\right)}{\sum\limits_{c'\in\mathcal{P}(i)}\exp\!\left(-\frac{\delta_{ic'}^2}{2\sigma^2}\right)}
}_{w_{ic}}
\mathbf{x}^{\downarrow}_c,
\label{eq:geo_up_impl}
\end{equation}
where \(\delta_{ic}\) is the spherical geodesic distance and \(\sigma\) is a fixed bandwidth.

\subsection{\emph{SO(3)}-Consistency Regularization}
\label{sec:reg}

\noindent\textbf{Removing absolute latitude encoding.}
To avoid re-introducing a privileged ``up'' direction, our main models do not inject absolute latitude positional encoding.
Positional reasoning is instead carried by quadrature-corrected local aggregation and the gauge-pooled relative bias.

\noindent\textbf{Index-based spherical resampling.}
We approximate spherical resampling under a 3D rotation \(g\in \emph{SO(3)}\) using nearest-neighbor index mappings
acting on (1) projected input tokens and (2) output logits.
Let \(\mathrm{idx}_{\mathrm{proj}}(g)\) be the index mapping applied to projected tokens, and
\(\mathrm{idx}_{\mathrm{img}}(g)\) the one applied to output logits. This is a discrete nearest-neighbor reindexing/resampling approximation on the sampled sphere, rather than an exact continuous \emph{SO(3)} action.

\noindent\textbf{Logit-space MSE consistency loss.}
Let \(f_{\mathrm{tok}}(\cdot)\) denote the branch that consumes projected tokens and returns final logits on the image/output sampling (i.e., the spherical-token backbone followed by the fixed output reprojection/head in our implementation).
Let \(\mathbf{x}_0\) denote projected tokens from the input panorama, and define the unrotated logits
\begin{equation}
\mathbf{z} = f_{\mathrm{tok}}(\mathbf{x}_0)\in\mathbb{R}^{L_{\mathrm{img}}\times C}.
\end{equation}
For a sampled rotation \(g\), we compute
\begin{equation}
\mathbf{z}_{\mathrm{rot}} = f_{\mathrm{tok}}\!\big(\mathbf{x}_0[\mathrm{idx}_{\mathrm{proj}}(g)]\big),
\qquad
\mathbf{z}_{\mathrm{tgt}} = \mathrm{stopgrad}\!\big(\mathbf{z}[\mathrm{idx}_{\mathrm{img}}(g)]\big),
\end{equation}
where \(\mathrm{idx}_{\mathrm{proj}}(g)\) and \(\mathrm{idx}_{\mathrm{img}}(g)\) are nearest-neighbor index mappings on the projected-token and image/output samplings, respectively, and \(\mathrm{stopgrad}(\cdot)\) blocks gradients through the target branch.
We minimize the mean-squared error in logit space (mean over all elements):
\begin{equation}
\mathcal{L}_{\mathrm{eq}} \;=\;
\frac{1}{L_{\mathrm{img}}C}\sum_{i=1}^{L_{\mathrm{img}}}\sum_{c=1}^{C}\left(
\mathbf{z}_{\mathrm{rot}}[i,c]-\mathbf{z}_{\mathrm{tgt}}[i,c]
\right)^2.
\label{eq:eq_mse_impl}
\end{equation}
The final training objective is
\begin{equation}
\mathcal{L} = \mathcal{L}_{\text{seg}} + \lambda\,\mathcal{L}_{\mathrm{eq}},
\end{equation}
where \(\mathcal{L}_{\text{seg}}\) is the standard per-node cross-entropy loss (with class-0 ignored in our setup).
This regularizer is used only during training and does not change the inference-time architecture.

\section{Experiments}
\label{sec:exper}

\subsection{Dataset and \emph{SO(3)} Evaluation Protocol}
\noindent\textbf{Pose35.}
All experiments are conducted on \textbf{Pose35}, a pose-perturbed variant of Stanford2D3D \cite{armeni2017joint2d3d}.
Pose35 is created by applying a \emph{deterministic per-sample} 3D rotation to each panorama using axis--angle uniform sampling
with a maximum rotation magnitude of \(35^\circ\) (fixed seed).
The same rotation is applied consistently to RGB, depth, and semantic labels via an inverse-map remapping of the equirectangular projection.
We use the standard Stanford2D3D train/val split, containing 999 training panoramas and 39 validation panoramas.

\noindent\textbf{Labels and metric.}
We evaluate 13 semantic classes.
In our pipeline, the \emph{unknown} label is encoded as class 0 and ignored when computing \emph{mIoU} (i.e., class 0 is excluded from the mean).
We report \emph{mIoU} as the primary metric.
For robustness, we report both the \emph{Base mIoU} on Pose35 val (no extra test-time rotation) and the \emph{SO(3) mIoU} under the stress test below.

\noindent\textbf{Input representation for spherical-token models.}
Following the SphereU\allowbreak Former family, the ERP panorama is first resampled to an icosphere-based image/output support (rank 7, vertex nodes).
For SO3UFormer, this sampled support is then projected to rank-6 spherical tokens for the internal U-shaped backbone, and final logits are reprojected back to rank-7 output sampling.
RGB is normalized with mean \(0.5\) and standard deviation \(0.225\).

\noindent\textbf{Training protocol.}
Unless otherwise specified by an ablation, we follow the same training protocol across SphereUFormer and SO3UFormer:
400 epochs, Adam optimizer with learning rate \(1\times 10^{-4}\), and distributed training on two GPUs with a global batch size of 32.
Data augmentation includes random yaw rotation and horizontal flip; color augmentation is disabled.
The segmentation objective is standard cross-entropy on logits with class-0 ignored.
When the \emph{SO(3)}-consistency regularizer is enabled, we sample one random 3D rotation per iteration (uniform quaternion),
apply the corresponding index-based resampling, and add the logit-space MSE penalty (Sec.~\ref{sec:reg}).

\noindent\textbf{SO(3) OOD stress test.}
To assess robustness beyond the training-time pose range, we evaluate models under \textbf{full 3D reorientations} that extend far outside Pose35's $\pm 35^\circ$ perturbations. Concretely, on the Pose35 validation set we additionally apply $10$ random $\mathrm{ZYX}$ Euler rotations, with yaw in $[0^\circ,360^\circ]$, pitch in $[0^\circ,180^\circ]$, and roll in $[0^\circ,360^\circ]$. We repeat the evaluation three times with a fixed random seed and report the mean \emph{mIoU} over all rotations and repeats. This is an \emph{out-of-distribution} robustness test: the underlying scene semantics are unchanged, while the camera frame is reoriented (up to representation-specific resampling effects). To ensure a fair comparison across representations, we use the same set of 3D rotations for all methods, while applying them in each method's native input domain---nearest-neighbor index mapping over icosphere normals for spherical-token models, and inverse-map ERP remapping for ERP-based baselines. This preserves a common geometric perturbation at the scene level without introducing an extra domain conversion that some baselines were not designed for. As a result, the reported \emph{SO(3)} mIoU primarily reflects robustness to camera-frame reorientation, while still capturing the practical discretization and resampling errors associated with each representation.

\subsection{Ablation Study}
\label{sec:ablation}
We ablate five design choices that progressively remove coordinate-frame shortcuts and improve \emph{SO(3)} robustness:
(1) removing absolute latitude encoding,
(2) quadrature-consistent attention (logit correction),
(3) gauge-pooled Fourier relative positional bias,
(4) geometry-consistent multi-scale sampling (area-weighted downsampling and geodesic-kernel upsampling),
and (5) the \emph{SO(3)}-consistency regularizer with weight \(\lambda\).
All variants are trained on Pose35 with the same optimization schedule and evaluated using the \emph{SO(3)} stress test above. Table~\ref{tab:ablation_pose35} reports the results.

% in preamble

\begin{table}[t]
\centering
\caption{Ablation on Pose35. \textbf{\emph{Base mIoU}}: Pose35 val without extra rotation.
\textbf{\emph{SO(3) mIoU}}: mean \emph{mIoU} under the \emph{SO(3)} OOD stress test. Best results are in bold.
For the last row, the \emph{SO(3)}-consistency regularizer is enabled with fixed weight \(\lambda=0.05\).}
\label{tab:ablation_pose35}
\small
\setlength{\tabcolsep}{3.5pt}
\renewcommand{\arraystretch}{1.15}
\begin{tabular}{c c c c c c c}
\hline
\makecell[c]{No abs.\\lat. PE} &
\makecell[c]{Quadrature\\attn.} &
\makecell[c]{Gauge-pooled\\bias} &
\makecell[c]{Geo.\\sampling} &
\makecell[c]{\(\mathcal{L}_{\mathrm{eq}}\)} &
\makecell[c]{\emph{Base}\\\emph{mIoU}} &
\makecell[c]{\emph{SO(3)}\\\emph{mIoU}} \\
\hline
- & - & - & - & -      & 67.53 & 25.26 \\
\cmark & - & - & - & -      & 68.64 & 64.66 \\
\cmark & \cmark & - & - & -      & 70.05 & 65.20 \\
\cmark & \cmark & \cmark & - & -      & 70.42 & 69.72 \\
\cmark & \cmark & \cmark & \cmark & -      & 70.92 & 69.90 \\
\cmark & \cmark & \cmark & \cmark & \cmark & \textbf{72.03} & \textbf{70.67} \\
\hline
\end{tabular}
\end{table}

\noindent\textbf{Findings.}
Removing absolute latitude encoding is the critical step for avoiding catastrophic failures under 3D reorientations, lifting \emph{SO(3) mIoU} from 25.26 to 64.66.
Quadrature-consistent attention further improves both base accuracy and robustness.
The gauge-pooled positional bias produces a substantial gain in \emph{SO(3)} stability, indicating that local tangent-plane angular geometry is more reliable than chart-dependent offsets.
Geometry-consistent sampling reduces the remaining robustness gap across scales.
Finally, the \emph{SO(3)}-consistency regularizer yields the best overall model, improving both \emph{Base} and \emph{SO(3)} \emph{mIoU} and resulting in the strongest and most stable configuration.

\subsection{Disentangling Architectural Robustness from Augmentation}
\label{sec:arch_vs_aug}
The main results train every model on Pose35, which entangles two possible sources of robustness: the architecture itself and exposure to rotated samples during training.
To isolate the architectural contribution, we retrain both SphereUFormer and SO3UFormer on the \emph{upright} Stanford2D3D dataset with no pose perturbation whatsoever, then evaluate at three increasing rotation levels. Table~\ref{tab:arch_vs_aug} reports the results.

\begin{table}[t]
\centering
\caption{Disentangling architectural robustness from rotation augmentation. Both models are trained on \emph{upright} Stanford2D3D with no pose perturbation, then evaluated on clean (upright) val, Pose35 ($\pm 35^\circ$ OOD), and Pose35 followed by the full $\mathrm{SO}(3)$ stress test.}
\label{tab:arch_vs_aug}
\setlength{\tabcolsep}{8pt}
\renewcommand{\arraystretch}{1.1}
\begin{tabular}{lccc}
\hline
Model (trained upright) & clean & Pose35 ($\pm 35^\circ$) & + full $\mathrm{SO}(3)$ \\
\hline
SphereUFormer \cite{benny2025sphereuformer} & 69.79 & 49.11 & 9.19 \\
SO3UFormer (ours) & 68.02 & \textbf{67.01} & \textbf{67.10} \\
\hline
\end{tabular}
\end{table}

\noindent\textbf{Analysis.}
The result is decisive.
Having never observed a single rotation during training, SO3UFormer degrades by less than one \emph{mIoU} from clean to full $\mathrm{SO}(3)$ ($68.02 \rightarrow 67.10$), whereas the gravity-anchored baseline collapses from $69.79$ to $9.19$, a loss exceeding sixty points.
The robustness of our design is therefore architectural, originating from the intrinsic operators themselves, and not a by-product of rotation augmentation.
We also note the small cost of breaking the gravity shortcut in the perfectly upright regime: our model trails the baseline by $1.77$ \emph{mIoU} on clean val, because gravity-aligned cues are genuinely predictive when the upright assumption holds.
This cost is recovered many times over under any reorientation, and it is eliminated entirely once pose-augmented training is permitted, where our model surpasses the baseline on upright accuracy as well (Table~\ref{tab:compare_pose35}).

\subsection{Comparison with State of the Art}
\label{sec:sota}
We compare SO3UFormer with recent panoramic segmentation baselines retrained on Pose35 using the same train/val split \cite{guttikonda2024single, benny2025sphereuformer,carlsson2024heal,ai2024elite360d}.
For each method, we follow its official training pipeline and evaluate with the identical \emph{SO(3)} stress test described above.
Table~\ref{tab:compare_pose35} summarizes the results.

\begin{table}[t]
\centering
\caption{Comparison on Pose35. All methods are retrained on Pose35 and evaluated with the identical $\mathrm{SO}(3)$ stress test. ``Type'' marks whether a method is non-equivariant (relying on coordinate cues and/or augmentation), strictly gauge-equivariant, or approximately equivariant. ``Gap'' is the absolute drop from \emph{Base} to \emph{SO(3)} \emph{mIoU} (lower is more robust). Best results are in bold.}
\label{tab:compare_pose35}
\setlength{\tabcolsep}{6pt}
\renewcommand{\arraystretch}{1.1}
\begin{tabular}{llcccc}
\hline
Method & Publication & Type & \emph{Base mIoU} & \emph{SO(3) mIoU} & Gap$\downarrow$ \\
\hline
SFSS \cite{guttikonda2024single} & WACV2024 & non-equiv. & 42.02 & 30.99 & 11.03 \\
HealSwin \cite{carlsson2024heal} & CVPR2024 & non-equiv. & 62.45 & 30.55 & 31.90 \\
Elite360 \cite{ai2024elite360d} & CVPR2024 & non-equiv. & 67.39 & 25.71 & 41.68 \\
SphereUFormer \cite{benny2025sphereuformer} & CVPR2025 & non-equiv. & 67.53 & 25.26 & 42.27 \\
IcoCNN \cite{cohen2019gauge} & ICML2019 & strict-equiv. & 59.94 & 59.17 & \textbf{0.77} \\
SO3UFormer (ours) & --- & approx.-equiv. & \textbf{72.03} & \textbf{70.67} & 1.36 \\
\hline
\end{tabular}
\end{table}

\noindent\textbf{Strict versus approximate equivariance.}
To anchor the equivariant end of the spectrum, we integrate IcoCNN~\cite{cohen2019gauge}, the most representative strictly gauge-equivariant model for spherical dense prediction, as a U-Net over the same rank-7 support under the identical Pose35 protocol.
It behaves exactly as its theory predicts: an $\mathrm{SO}(3)$ gap of only $0.77$ \emph{mIoU}, the smallest of any method, but at a pronounced cost in capacity (\emph{Base mIoU} $59.94$, trailing ours by $12.1$ points), a textbook instance of the capacity--equivariance trade-off~\cite{cohen2018sphericalcnns,weiler2019e2cnn}.
The two failure modes thus bracket our method: gravity-anchored Transformers are accurate upright but collapse under reorientation (gap $>40$), while strictly equivariant convolutions are robust but capacity-limited.
SO3UFormer is Pareto-superior to both, leading in \emph{Base} and \emph{SO(3)} \emph{mIoU} at a near-equivariant gap of $1.36$, which substantiates our thesis that approximate equivariance through intrinsic locality preserves capacity while recovering most of the robustness.

\noindent\textbf{Discussion.}
While several baselines achieve reasonable \emph{Base mIoU} on Pose35, their performance collapses under full \emph{SO(3)} reorientations.
In contrast, SO3\allowbreak UFormer preserves accuracy under the same stress test, narrowing the gap between Base and \emph{SO(3)} performance and establishing a new state of the art for rotation-robust panoramic segmentation.

\subsection{Axis-Wise Robustness: Why \texorpdfstring{$\mathrm{SO}(3)$}{SO(3)}, Not \texorpdfstring{$\mathrm{SO}(2)$}{SO(2)}}
\label{sec:axis}
A natural question is whether panoramic robustness really requires the full rotation group $\mathrm{SO}(3)$, or only the one-parameter subgroup $\mathrm{SO}(2)$ of rotations about the gravity axis.
To answer this empirically, we decompose the stress test into three single-axis families, yaw (about gravity, $z$), pitch ($x$), and roll ($y$), and contrast them with the full joint group.
Table~\ref{tab:axis_breakdown} reports the result.

\begin{table}[t]
\centering
\caption{Axis-wise robustness on Pose35 val. The $\mathrm{SO}(3)$ stress test is decomposed into single-axis rotations and the full joint group, averaged over the same magnitudes and three seeds as the main protocol.}
\label{tab:axis_breakdown}
\setlength{\tabcolsep}{8pt}
\renewcommand{\arraystretch}{1.1}
\begin{tabular}{lcccc}
\hline
Method & yaw-only & pitch-only & roll-only & full $\mathrm{SO}(3)$ \\
\hline
SphereUFormer \cite{benny2025sphereuformer} & 67.06 & 34.80 & 30.32 & 25.26 \\
SO3UFormer (ours) & \textbf{71.09} & \textbf{70.76} & \textbf{70.96} & \textbf{70.67} \\
\hline
\end{tabular}
\end{table}

\noindent\textbf{Analysis.}
Rotations about the gravity axis are effectively trivial under the equirectangular representation: a yaw is a horizontal circular shift of the panorama and is already covered by standard yaw augmentation, so the baseline retains $67.06$ \emph{mIoU} under yaw-only perturbations, close to its base accuracy on Pose35.
The catastrophic degradation appears only once the rotation leaves the gravity axis: the same baseline drops to $34.80$ under pitch and $30.32$ under roll, and to $25.26$ under the joint group.
In other words, the two additional degrees of freedom beyond yaw, precisely the ones absent from $\mathrm{SO}(2)$, are what define the difficulty of the task.
Unconstrained capture in practice (a banking drone, a jittering handheld rig, a robot gripper) couples all three axes simultaneously, so the operative nuisance group is $\mathrm{SO}(3)$ rather than $\mathrm{SO}(2)$.
SO3UFormer is uniformly stable across all four families, confirming that its robustness is not specific to any single rotation axis.

\subsection{Cross-Dataset Generalization: Matterport3D}
\label{sec:mp3d}
All experiments so far are confined to Stanford2D3D. A natural concern is whether the observed collapse-versus-stability behavior is specific to that dataset and its 13-class label set. To test this, we repeat the core experiment on a second, independently captured real-world indoor dataset with a different label taxonomy.

\noindent\textbf{Setup.}
We use 360FV-Matterport, the front-view panoramic split of Matterport3D~\cite{chang2017matterport3d} released by 360BEV~\cite{teng2024_360bev}, comprising equirectangular panoramas with 20-class semantic labels under the official 61/7/18 building split (7{,}829 train, 772 validation panoramas). Raw MPcat40 indices are mapped to the 20 classes following~\cite{teng2024_360bev}, with unlabeled pixels ignored. Both SphereUFormer and SO3UFormer are trained from scratch on \emph{upright} Matterport3D with no pose perturbation, using the same optimizer, resolution, and node sampling as the Stanford2D3D experiments (80 epochs, matched to the larger training set). This mirrors the upright-training protocol of Sec.~\ref{sec:arch_vs_aug}, isolating architectural robustness from pose augmentation. We report clean validation mIoU, the identical full $\mathrm{SO}(3)$ stress test (10 rotations, 3 repeats, seed 123), and the single-axis decomposition of Sec.~\ref{sec:axis}. Table~\ref{tab:mp3d} summarizes the results.

\begin{table}[t]
\centering
\caption{Cross-dataset generalization on 360FV-Matterport~\cite{teng2024_360bev} (20 classes). Both models are trained from scratch on \emph{upright} Matterport3D and evaluated under the same single-axis and full $\mathrm{SO}(3)$ protocol as Table~\ref{tab:axis_breakdown}. The baseline reproduces the exact collapse pattern (yaw-invariant, destroyed by pitch and roll), whereas SO3UFormer is essentially rotation-invariant. Gap is clean minus full $\mathrm{SO}(3)$.}
\label{tab:mp3d}
\setlength{\tabcolsep}{6pt}
\renewcommand{\arraystretch}{1.1}
\begin{tabular}{lcccccc}
\hline
Method & clean & yaw & pitch & roll & full $\mathrm{SO}(3)$ & Gap$\downarrow$ \\
\hline
SphereUFormer \cite{benny2025sphereuformer} & 27.48 & 27.54 & 8.99 & 7.32 & 3.84 & 23.64 \\
SO3UFormer (ours) & 24.65 & \textbf{24.81} & \textbf{24.70} & \textbf{24.84} & \textbf{24.68} & \textbf{-0.03} \\
\hline
\end{tabular}
\end{table}

\noindent\textbf{Analysis.}
The phenomenon transfers cleanly to the new dataset. The gravity-anchored baseline again collapses under full $\mathrm{SO}(3)$, from $27.48$ to $3.84$ mIoU (gap $23.64$), and the single-axis breakdown reproduces the Stanford2D3D signature exactly: yaw is harmless ($27.54$, absorbed by the equirectangular wrap), while pitch and roll are catastrophic ($8.99$ and $7.32$). SO3UFormer, by contrast, is essentially rotation-invariant on a second dataset, moving only from $24.65$ to $24.68$ under full $\mathrm{SO}(3)$ (gap $-0.03$) and staying within $0.2$ mIoU across every single-axis family. Two points deserve emphasis. First, having never seen a rotation during training, SO3UFormer degrades by effectively zero, confirming that its robustness is architectural, exactly as on Stanford2D3D (Sec.~\ref{sec:arch_vs_aug}). Second, the small clean deficit relative to the baseline ($2.83$ mIoU) is the same price of removing the gravity shortcut documented there ($1.77$ mIoU on upright-trained Stanford2D3D), and it is dwarfed by the robustness it buys. The absolute mIoU is lower than methods built on ImageNet-pretrained backbones, because our spherical models are trained from scratch; the experiment isolates rotation robustness, for which the base-to-$\mathrm{SO}(3)$ gap, not the absolute score, is the relevant quantity.

\subsection{Cross-Task Generalization: Panoramic Depth Estimation}
\label{sec:depth}
The second generalization question is whether the intrinsic design is specific to semantic segmentation or confers rotation robustness on dense spherical prediction in general. We therefore apply the same architecture to monocular panoramic depth estimation on Pose35, changing nothing but the output head (the 13-class classifier is replaced by a single-channel regressor) and the loss (a \texttt{BerhuLoss} depth objective). Both models are trained from scratch for 400 epochs under the same optimization and augmentation protocol as the main segmentation experiments. Depth is evaluated over pixels with valid ground truth ($0 < d \le 10$\,m) using AbsRel ($\downarrow$), RMSE ($\downarrow$), and $\delta_1$ ($\uparrow$, the fraction of pixels with $\max(d/\hat{d},\hat{d}/d)<1.25$), and under the identical $\mathrm{SO}(3)$ stress test. Table~\ref{tab:depth} reports these results.

\begin{table}[t]
\centering
\caption{Cross-task generalization to panoramic depth estimation on Pose35 under the same $\mathrm{SO}(3)$ stress test used for segmentation. The gravity-anchored baseline collapses ($\delta_1$: $0.852 \to 0.316$), whereas SO3UFormer degrades by under $1\%$ on every metric.}
\label{tab:depth}
\setlength{\tabcolsep}{5pt}
\renewcommand{\arraystretch}{1.1}
\begin{tabular}{l ccc ccc}
\hline
 & \multicolumn{3}{c}{Base (Pose35 val)} & \multicolumn{3}{c}{full $\mathrm{SO}(3)$ OOD} \\
\cmidrule(lr){2-4}\cmidrule(lr){5-7}
Model & AbsRel$\downarrow$ & RMSE$\downarrow$ & $\delta_1\uparrow$ & AbsRel$\downarrow$ & RMSE$\downarrow$ & $\delta_1\uparrow$ \\
\hline
SphereUFormer \cite{benny2025sphereuformer} & 0.134 & 0.540 & 0.852 & 0.437 & 1.531 & 0.316 \\
SO3UFormer (ours) & \textbf{0.118} & \textbf{0.488} & \textbf{0.879} & \textbf{0.119} & \textbf{0.489} & \textbf{0.877} \\
\hline
\end{tabular}
\end{table}

\noindent\textbf{Analysis.}
The pattern mirrors segmentation but is, if anything, more extreme. Under full $\mathrm{SO}(3)$ reorientation the baseline's AbsRel inflates by $226\%$ ($0.134 \to 0.437$) and $\delta_1$ collapses from $0.852$ to $0.316$: once the camera frame leaves the training distribution, the network can no longer recover scene geometry. SO3UFormer degrades by less than $1\%$ on every metric (AbsRel $0.118 \to 0.119$; $\delta_1$ $0.879 \to 0.877$) and, because this experiment follows the pose-augmented protocol of the main results, it also surpasses the baseline in the base setting. The near-zero degradation confirms that the intrinsic operators of Sec.~\ref{sec:metho} are not segmentation-specific: they address a task-agnostic cause, the coupling of absolute coordinates with measure-inconsistent aggregation, and therefore transfer to dense spherical regression as well.

\subsection{Computational Cost}
\label{sec:efficiency}
A concern for any geometric attention mechanism is whether the added structure, here the per-pair plane-slice projection, geodesic distances, and the sum over six in-plane rotations and $F$ anchors, inflates inference cost.
We measure inference on a single GPU with one rank-7 icosphere panorama in evaluation mode; Table~\ref{tab:efficiency} reports these measurements.

\begin{table}[t]
\centering
\caption{Inference cost on a single GPU with one rank-7 icosphere panorama, in evaluation mode. MACs are measured with \texttt{fvcore}; FPS is the median over 100 forward passes after 10 warmup passes. The static gauge-pooled bias and the quadrature term ($\log\omega_j$) are materialized as inference buffers, and the training-only consistency loss is excluded from the inference path.}
\label{tab:efficiency}
\setlength{\tabcolsep}{8pt}
\renewcommand{\arraystretch}{1.1}
\begin{tabular}{lcccc}
\hline
Model & Params (M) & MACs (G) & FPS & Peak Mem.\ (MB) \\
\hline
SphereUFormer \cite{benny2025sphereuformer} & 14.92 & 14.92 & 7.58 & 912 \\
SO3UFormer (ours) & \textbf{14.57} & \textbf{13.62} & \textbf{9.36} & \textbf{904} \\
\hline
\end{tabular}
\end{table}

\noindent\textbf{Analysis.}
The apparent expense of the gauge-pooled bias is amortized away at inference: every angular quantity depends only on the fixed mesh and is precomputed as a buffer, so each query--key pair costs a single bilinear lookup rather than a recomputed six-rotation, $F$-anchor sum. SO3UFormer is therefore marginally \emph{cheaper} than the baseline on every axis, with fewer parameters and MACs, $23\%$ higher throughput, and lower peak memory: intrinsic geometric reasoning need not compromise deployability on real-time mobile or aerial platforms.

\subsection{Qualitative Visualization}
\label{sec:qual}

\begin{figure*}[t]
    \centering
    \includegraphics[width=\textwidth]{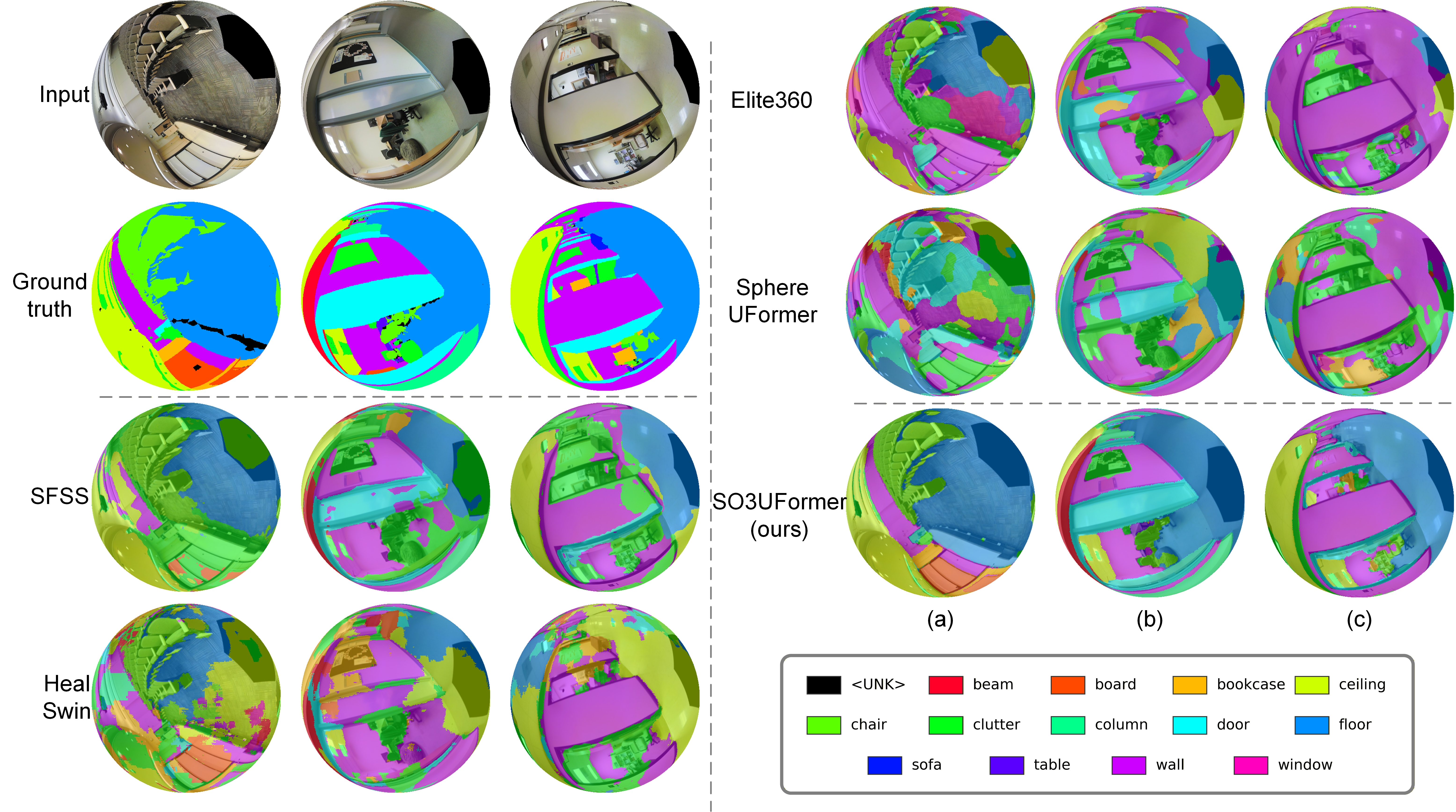}
    \caption{\textbf{Qualitative comparison under the out-of-distribution \emph{SO(3)} stress test on Pose35 validation.}
    Three representative scenes are shown (a)--(c), each evaluated under arbitrary 3D reorientation.
    For compact presentation, the methods are arranged across the left and right halves of the figure, but all predictions correspond to the same rotated inputs.
    Compared with SFSS, HealSwin, Elite360, and SphereUFormer, our \textbf{SO3UFormer} produces substantially more stable and semantically coherent layouts under full \emph{SO(3)} perturbations, with reduced large-scale label drift and structural inconsistencies.
    The color legend at the bottom shows the 13-class palette used for visualization (the unknown class is excluded from \emph{mIoU} evaluation).}
    \label{fig:qualitative}
\end{figure*}

Figure~\ref{fig:qualitative} presents qualitative results on Pose35 validation under the \emph{SO(3)} out-of-distribution stress test.
We show three rotated examples and compare four baselines (SFSS, HealSwin, Elite360, and SphereUFormer) with SO3UFormer.

A clear pattern appears across all three scenes. After 3D reorientation, several competing methods fail on categories that are usually easy to recognize in upright panoramas, especially \emph{floor}. In many cases, HealSwin, Elite360, and SphereUFormer misclassify large floor regions as \emph{ceiling} or \emph{wall}. This behavior is consistent with a gravity-dependent bias: the models implicitly expect floor-like semantics to appear near the bottom of the panorama, and that shortcut breaks once the camera orientation changes. SFSS is relatively more stable on the floor class in some examples, but it still shows noticeable boundary errors and local inconsistencies.

Another visible failure mode of the baselines is the presence of coarse, blocky, or jagged prediction patterns, particularly around object boundaries and thin structures. These artifacts are clearly visible in the \emph{SO(3)} OOD stress test and indicate poor generalization when the test orientation departs substantially from the gravity-aligned regime seen during training. In contrast, SO3UFormer remains much more stable under the same rotations. It recovers the correct semantic categories more consistently, including large floor and wall regions, and it preserves sharper transitions at boundaries. The qualitative results match the quantitative findings in Table~\ref{tab:compare_pose35}: reducing dependence on the global gravity axis is essential for reliable panoramic segmentation under unconstrained camera motion.

\section{Discussion and Limitations}
\label{sec:discussion}

\noindent\textbf{Where the gains come from.}
Two experiments jointly localize the source of robustness. The ablation (Table~\ref{tab:ablation_pose35}) shows that removing the absolute-latitude encoding accounts for the largest single jump, since it is the only operator explicitly tied to a global axis, while the remaining intrinsic operators add smaller but consistent gains that also improve upright accuracy, because measure- and gauge-consistent local processing is simply a better-posed description of spherical geometry regardless of orientation. The upright-training study (Table~\ref{tab:arch_vs_aug}) then confirms that this robustness is architectural rather than a by-product of pose augmentation; its only real cost, the small upright deficit from discarding the gravity cue, is quantified in Sec.~\ref{sec:arch_vs_aug} and is erased once pose-augmented training is allowed.

\noindent\textbf{Approximate, not exact, equivariance.}
As formalized in Sec.~\ref{sec:metho}, SO3UFormer is equivariant only in the continuous limit; the icosahedral discretization and nearest-neighbor resampling introduce a residual gap that the consistency regularizer suppresses but does not eliminate.
A strictly gauge-equivariant network such as IcoCNN closes this gap almost completely (Table~\ref{tab:compare_pose35}) at a substantial cost in base accuracy.
Bridging the two regimes, for instance through higher-fidelity equivariant resampling or steerable attention kernels, is an open direction.

\noindent\textbf{Scope of evaluation.}
Our study isolates camera-frame reorientation on indoor panoramas, where strong vertical regularities (floor, ceiling, horizon) make the gravity shortcut maximally tempting and therefore make its removal maximally informative.
The $\mathrm{SO}(3)$ perturbations are applied through representation-native resampling rather than physical re-capture, so they model the geometric effect of reorientation but not sensor-specific artifacts such as motion blur or rolling-shutter distortion.
Extending the protocol to outdoor panoramas and to datasets with real tracked attitude (handheld or drone-mounted rigs) would test the approach under genuine attitude variation; to our knowledge, no public outdoor panoramic semantic-segmentation benchmark with a comparable label protocol currently exists, which is itself a gap worth closing.

\section{Conclusion}
\label{sec:concl}

We identify a practical but under-evaluated failure mode in panoramic dense prediction: models trained under gravity-aligned assumptions do not reliably generalize to unconstrained 3D reorientation. To address this, we introduced \textbf{SO3UFormer}, which improves rotation robustness by removing absolute latitude bias and combining quadrature-consistent local attention, gauge-pooled angular relative bias, geometry-consistent multi-scale sampling, and a logit-space \emph{SO(3)}-consistency regularizer. On Pose35 and a full \emph{SO(3)} OOD stress test, SO3UFormer substantially narrows the gap between upright and rotated performance and exceeds recent panoramic baselines, underscoring the value of geometry-consistent positional reasoning and multi-scale operators. The same behavior reproduces on a second real-world dataset (Matterport3D) and on panoramic depth estimation, indicating that the robustness is a property of the intrinsic formulation rather than of a single benchmark or task.

Future work should move beyond controlled rotations to datasets with real attitude variation (e.g., indoor drones or handheld rigs with tracked 6-DoF pose), potentially supported by photorealistic simulation for scale. Extending the same geometric design to other spherical dense prediction tasks (such as depth) and testing transfer to broader panoramic architectures (e.g., detection and tracking under large viewpoint changes) are also promising directions. Higher-fidelity spherical resampling for training-time consistency remains an open avenue.

\bmhead{Acknowledgements}

This work was supported by the Xi'an Jiaotong-Liverpool University Postgraduate
Research Scholarship under Grant No. FOS2210JJ03.

\bmhead{Data Availability}

The Stanford2D3D and Matterport3D datasets used in this study are publicly available from their respective official sources, subject to the applicable terms of use. The Pose35 benchmark was generated from Stanford2D3D using the rotation protocol described in this paper. The source code for generating Pose35, together with the implementation, trained models, and evaluation scripts, is publicly available in our GitHub repository. No new raw data were collected for this study.

\bibliography{sn-bibliography}% common bib file

\end{document}